\documentclass[a4paper, 10pt, conference]{ieeeconf} 
\AtBeginDocument{\let\autocite\cite}

\IEEEoverridecommandlockouts                              

\usepackage{balance}
\usepackage{url}
\usepackage{cite}
\usepackage{graphicx} 
\usepackage{flushend}
\usepackage{booktabs}
\usepackage{tikz}
\usepackage{tabularx}
\newcommand\copyrighttext{%
  \footnotesize \textcopyright 2026 IEEE.
  Permission from IEEE must be obtained for all uses, in any current or future
  media, including reprinting/republishing this material for advertising or promotional
  purposes, creating new collective works, for resale or redistribution to servers or
  lists, or reuse of any copyrighted component of this work in other works.}
\newcommand\copyrightnotice{%
\begin{tikzpicture}[remember picture,overlay]
\node[anchor=south,yshift=10pt] at (current page.south) {\fbox{\parbox{\dimexpr\textwidth-\fboxsep-\fboxrule\relax}{\copyrighttext}}};
\end{tikzpicture}%
}

\begin{document}
\IEEEoverridecommandlockouts
\overrideIEEEmargins

\title{\LARGE \bf
  Toward Self-Repairing Ubiquitous Robots Using Goal-Oriented Agentic AI in Human-Robot Interactions
}

\author{Morten Roed Frederiksen$^{1}$
  \thanks{{$^{1}$Morten Roed Frederiksen {\tt\small mrof@itu.dk} is affiliated with the Data Systems and Robotics (DARSA) Department of The IT-University of Copenhagen Denmark.}}
}

\maketitle
\copyrightnotice
\begin{abstract}
Ubiquitous robotic systems often lack traditional visual interfaces, making natural language interaction important for maintenance and repair. This paper presents a goal-oriented agentic AI architecture that enables non-expert users to complete technical repair tasks through situated dialogue. The architecture separates pre-interaction goal decomposition, persistent state tracking, strategic goal management, and real-time conversational execution. We evaluated the system in a physical hardware repair task with twenty participants. Nineteen participants completed the task, corresponding to a 95\% completion rate. Participants rated the system as helpful and competent, and the agent remained robust to conversational diversions such as meta-queries and code-switching. A comparison with a prior online baseline showed that physical interaction significantly reduced perceived social presence, \(p=.0005\), and trust and competence, \(p=.037\), while perceived helpfulness remained high.
\end{abstract}

\section{Introduction}
Ubiquitous robotic systems worn on or close to the body are exposed to movement, pressure, perspiration, and everyday handling \cite{Zhao02102023, chen23, s22197584, Heikenfeld2018WearableSM}. These conditions make maintenance and repair important for long-term use, yet user-led repair remains difficult because robotic systems often contain hidden dependencies, branching procedures, and technical concepts that exceed a non-expert user's working memory \cite{Sweller1988CognitiveLD, Hancock07, Kiesler2002MentalMO}. Static manuals and linear tutorials are poorly suited to such tasks because repair often requires conditional instructions, clarification, and recovery from partial progress or mistakes \cite{buttussi21, Henderson2011ExploringTB}.

This paper investigates goal-oriented agentic AI as an interaction mechanism for ubiquitous robot repair. Rather than presenting a fixed troubleshooting script, the proposed architecture converts natural-language repair requirements into structured goals, tracks user progress over time, and uses a conversational agent to guide the user through the currently active objective. This is particularly relevant for wearable and concealed devices, where visual feedback may be unavailable or socially inappropriate, and where speech can support eyes-free interaction during physical manipulation \cite{10.1145/191666.191732, Brewster2003MultimodalI, Profita2013DontMM, chulhong2015, Marge2020SpokenLI}. We use the term self-repairing to refer to robot-initiated,
user-mediated repair, where the robot guides a human through the physical intervention rather than executing the repair autonomously.

The design challenge is that repair dialogue must be both flexible and constrained. A system that is too script-like cannot answer clarification questions or recover from unexpected user behavior. A system that is too open-ended may lose track of the repair state, skip required checks, or provide instructions that are inconsistent with the user's current physical configuration. We therefore treat repair as a goal-management problem: the agent should be conversationally responsive while remaining grounded in a persistent representation of the task state and its conditional dependencies.

Prior work on speech interfaces has shown the value of constraining interaction to the current task state, while recent work on LLM-based agents demonstrates how reasoning and acting can be combined for goal-directed behavior \cite{794797, Yao2022ReActSR, Ahn2022DoAI}. Our work adapts this idea to a situated repair context by separating high-level goal reasoning from low-latency conversational execution. This separation allows the robot to maintain a repair plan while still responding naturally to user questions, clarifications, and diversions.

We evaluate the architecture in a human-robot interaction study where participants repaired a physical hardware puzzle through spoken dialogue with an animated robot interface. The task required participants to identify a color-coded condition, connect the correct wires, and verify a status lamp. We additionally compare the results with a prior online version of the same goal structure, allowing us to examine whether the same agentic logic is experienced differently when users manipulate physical hardware rather than a virtual interface.

The results show that 19 of 20 participants completed the repair, while the comparison with the online baseline indicates that the physical setting lowered perceived social presence and trust despite preserving high helpfulness. The contributions of this paper are: (1) a goal-oriented multi-agent architecture for non-linear repair dialogue; (2) a physical evaluation of the architecture in an eyes-free ubiquitous robotics scenario; and (3) evidence that situated physical repair changes user perception relative to an online version of the same task.

\section{Methods}
\subsection{System Architecture}
\begin{figure}[h]
\centering
\includegraphics[width=0.48\textwidth]{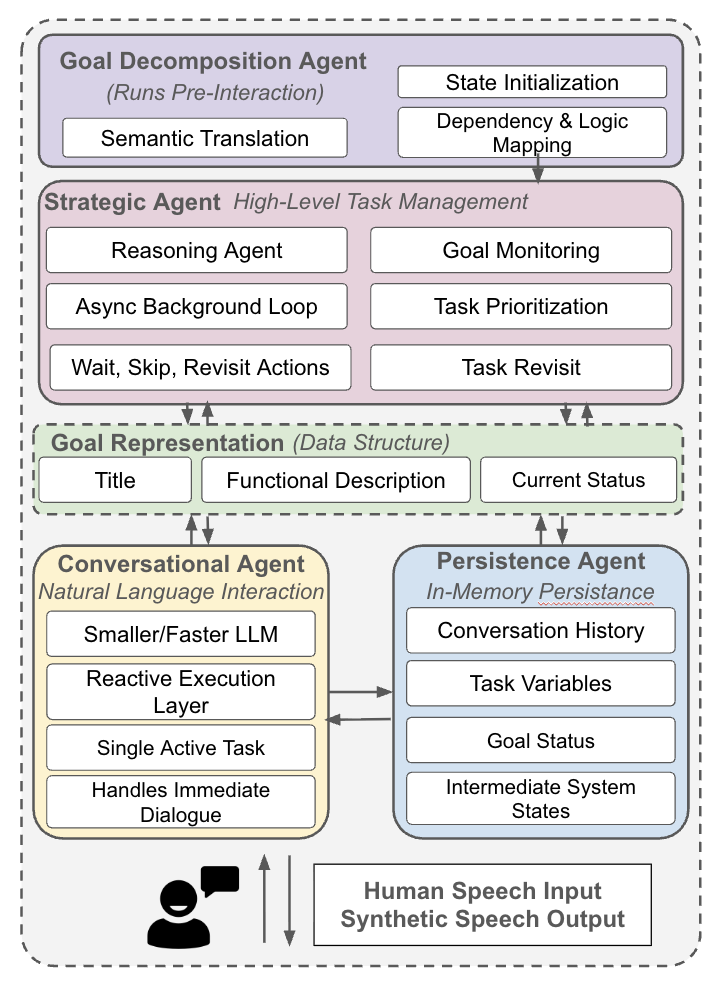}
\caption{Goal-oriented multi-agent architecture. The goal decomposition agent creates the initial goal hierarchy before interaction. The persistence agent stores conversation history, task variables, goal status, and intermediate states. The strategic agent selects the active goal, and the conversational agent produces real-time user-facing responses.}
\label{fig:agentic_architecture}
\end{figure}

We developed a goal-oriented agentic AI framework for non-linear conversational repair tasks. The architecture contains four functional units: a \textit{Goal Decomposition Agent}, a \textit{Persistence Agent}, a \textit{Strategic Agent}, and a \textit{Conversational Agent} (Figure \ref{fig:agentic_architecture}). Together, these components separate offline task structuring, state persistence, high-level goal selection, and real-time dialogue.

The \textit{Goal Decomposition Agent} converts unconstrained repair instructions into a hierarchy of goals and subgoals before the live interaction. This offline step was implemented with Llama 3.1 70B and encoded the conditional structure of the task: participants first reported whether the sticker was blue or green; the required wire connections then depended on that color; finally, the participant verified that the confirmation lamp was lit. Each goal was represented by a title, description, completion condition, and status, allowing the system to track whether goals were pending, active, completed, or skipped. This representation made it possible to preserve the structure of a repair manual without forcing the user through a strictly linear script.

The \textit{Persistence Agent} stored the evolving interaction state, including conversation history, task variables, and intermediate goal states, using a ChromaDB database \cite{chromadb_github}. This enabled the system to maintain coherence across turns, avoid redundant instructions, and recover when users asked clarification questions or temporarily moved away from the main task. Persistence was particularly important in the physical condition because the user's spoken progress and the state of the hardware could diverge: a participant might ask for a repeated instruction, connect only part of the wiring sequence, or return to an earlier step after handling the clips.

The \textit{Strategic Agent} governed high-level goal management. It used the persisted state to determine which repair objective should be active and whether the user had satisfied the current goal. This layer enforced the conditional task logic while allowing non-linear dialogue and revisits to prior objectives when needed. The \textit{Conversational Agent} was the user-facing component, implemented with Qwen 30B to support lower-latency speech interaction. It generated responses only for the active goal selected by the strategic layer and signaled goal completion back to the planner rather than advancing the workflow independently. This design prevented the conversational layer from skipping ahead when users engaged in chit-chat or asked out-of-scope questions.

\subsection{Experimental Procedure and Testbed}
The system was evaluated with 20 university participants (12 male, 7 female, 1 non-binary), aged 18--44. Participants interacted with an animated robot face implemented in Pygame. The graphical interface ran on a MacBook Pro, while LLM inference ran on a secondary server. The robot used face detection to initiate engagement, microphone thresholding and Whisper-based speech-to-text for user input, and text-to-speech for verbal responses. To support social framing, the animated face exhibited simple autonomous behaviors such as blinking, smiling, and gaze shifts.

Participants repaired a hardware testbed consisting of a wooden box with a green status lamp, ten numbered wires with crocodile clips, and a blue or green sticker indicating the task variation. To complete the task, participants had to tell the robot the sticker color, follow the corresponding wiring instructions, and confirm that the lamp was active. The internal circuitry lit the lamp only when the correct four wires were connected. The task therefore combined a simple diagnostic condition with a physical manipulation sequence and an objective completion signal. This setup was intentionally simple but embodied the central structure of many repair tasks: an initial diagnostic observation, a conditional action sequence, and an observable confirmation state.

The procedure included a pre-study questionnaire on demographics and prior familiarity with robotics and AI chatbots, a two-minute warm-up conversation with the robot, the repair task guided by the agentic architecture, and a post-study questionnaire on perceived interaction quality and pedagogical effectiveness. Objective interaction data, including completion, turn counts, message lengths, and session duration, were logged and merged with the questionnaire responses. One objective interaction log was missing and excluded from objective turn-count analysis.

To contextualize the physical study, we compared the results with a prior online baseline \cite{frederiksen2026goal}. The baseline used the same goal-oriented architecture and branching repair logic, but participants completed a virtual wiring task rather than manipulating physical wires. Holding the task logic constant allowed us to examine how situated physical interaction affected task performance and user perception.

\section{Results}
\subsection{Task Completion and User Perception}
Nineteen of the twenty participants completed the physical repair task, corresponding to a 95\% completion rate. Completion was verified through successful termination of the repair sequence and confirmation that the status lamp was active. This indicates that the agentic architecture reliably guided most users through a multi-step, conditionally branched physical repair task.

Post-interaction questionnaire responses showed that participants rated the system as helpful and competent. The Trust and Competence composite was above the neutral midpoint of the 7-point scale ($M=4.87, SD=1.57, \alpha=.93, d=0.56$), while Social Presence was close to neutral ($M=4.09, SD=1.45, \alpha=.80, d=0.06$). Participants also reported positive perceived self-efficacy after the task ($M=4.95$), suggesting that the interaction did not merely produce completion, but also supported users' sense that they could carry out the repair. Prior AI familiarity was not significantly correlated with Trust and Competence ($r=-.07, p=.775$), suggesting that evaluations were driven more by the interaction itself than by pre-existing attitudes toward AI.

Interaction logs showed that participants used more turns than the theoretical optimal path of 9 turns. The mean completion length was 19.7 turns ($SD=20.6$), significantly above the optimum ($t(19)=2.27, p=.035$). This variability reflected clarification requests, social diversions, and the added demands of manipulating physical crocodile clips. The system nevertheless recovered from non-normative inputs: code-switching occurred in 10.5\% of sessions, and meta-queries or chit-chat occurred in 21.1\% of sessions. In these cases, the agent either mirrored the user's language or briefly handled the diversion before redirecting the interaction to the repair goal.

\begin{table}[h]
\centering
\caption{Key outcomes and comparison with online baseline.}
\label{tab:short_results}
\scriptsize
\begin{tabular}{lccc}
\toprule
\textbf{Metric} & \textbf{Online} & \textbf{Physical} & \textbf{Result} \\
\midrule
Task completion & 100\% & 95\% & High in both \\
Conversational turns & 11.4 & 19.7 & $p=.095$ \\
System helpfulness & 4.21 & 4.60 & $p=.37$ \\
Perceived understanding & 4.21 & 4.25 & $p=.89$ \\
Social presence & 5.55 & 4.09 & $p=.0005$ \\
Trust \& competence & 5.74 & 4.87 & $p=.037$ \\
\bottomrule
\end{tabular}
\end{table}

\subsection{Comparison with Online Baseline}
The physical study was compared with a prior online baseline using the same architecture and branching repair logic. Helpfulness remained high in the physical setting ($M=4.60$) and was not significantly different from the online baseline ($M=4.21, p=.37$). Perceived understanding was also stable ($M=4.25$ physical vs. $M=4.21$ online, $p=.89$). This suggests that the architecture's instructional value transferred from the virtual setting to the physical setting, even though the physical task required users to manipulate wires, remember spoken instructions, and verify the visible lamp state.

However, Social Presence decreased from $M = 5.55$ in the online baseline to $M = 4.09$ in the physical study, and Trust and Competence decreased from $M=5.74$ to $M=4.87$ ($p=.037$). Conversational turns increased from 11.4 to 19.7, but this difference was not significant ($p=.095$). Together, these results indicate that physical situatedness did not undermine perceived usefulness, but it did make the interaction feel less socially compelling and somewhat less trustworthy than the online version.

\section{Discussion and Conclusion}
The results suggest that goal-oriented agentic AI can support user-led repair of ubiquitous robotic systems. The 95\% completion rate demonstrates that the architecture could maintain task progress in a physical setting where users had to manipulate hardware, ask clarifying questions, and occasionally diverge from the main task. The separation between strategic goal management and conversational execution appears useful for this context: the system could respond fluidly while remaining anchored to the repair procedure.

The findings also show why a purely conversational agent is not sufficient for repair. In an open-ended repair setting, the system must remember what the user has already reported, maintain the current goal, determine whether the task state has changed, and decide when to move forward. By storing goal status and routing the dialogue through the active objective, the architecture provided a middle ground between a static manual and an unconstrained chatbot. This is particularly valuable for ubiquitous robots because the relevant components may be hidden, screen-based feedback may be limited, and the user's hands may be occupied.

At the same time, the comparison with the online baseline shows that physical repair changes how users evaluate the agent. The lower ratings for social presence and trust may reflect a reality gap between virtual and physical interaction. In the online version, connecting wires was a low-stakes digital action, whereas the physical task required participants to manipulate real connectors and potentially worry about making mistakes. Even when the task was safe, the tangible consequences of hardware manipulation may have made participants more conservative in judging the system's competence. Thus, physical embodiment does not automatically increase social evaluation; in this case, it increased the practical demands placed on the agent.
This functional-social split is important for the design of repair-oriented agents. The physical system preserved perceived usefulness and produced high task completion, but users evaluated the agent more cautiously when its instructions had to be enacted on hardware. This suggests that situated repair agents should not only provide correct procedural guidance, but also actively manage the user’s confidence in the physical consequences of each step. In future robots, this could involve explicit safety statements, step-by-step progress markers, confirmation prompts before irreversible actions, or sensor-based verification that allows the robot to acknowledge what the user has physically done.

For designers of ubiquitous robotic systems, the results point to two implications. First, conversational repair systems should expose progress and confirmation cues whenever possible, because users may need reassurance when acting on physical hardware. Second, goal-oriented state tracking should be treated as part of the interaction design rather than as a backend implementation detail. The user's experience of trust depends not only on what the agent says, but also on whether it consistently remembers the repair context and recovers from deviations.
Rather than demonstrating fully autonomous diagnosis, this study demonstrates a narrower but important capability: maintaining a coherent repair dialogue while a non-expert user performs the physical actions. This distinction matters for ubiquitous robots because many near-term repair scenarios may not require the robot to sense every hardware state directly. Instead, the system must combine user-reported observations, persistent task state, and goal-directed dialogue to help users act safely and sequentially.

This study has limitations. The physical evaluation involved only twenty participants and a constrained wiring task, so the results should be interpreted as feasibility evidence rather than a definitive evaluation across repair contexts. The online baseline also differed in sample size and interaction medium, which limits causal claims about embodiment. Future work should test the architecture on longer repair procedures, richer diagnostic uncertainty, repeated interactions over time, and tasks where the system can sense more of the hardware state directly. It should also investigate adaptive goal strictness, where the system dynamically balances task focus with more open conversational behavior.

In summary, this paper presented a goal-oriented agentic AI architecture for situated repair dialogue. By decoupling goal decomposition, persistence, strategic planning, and real-time conversation, the system enabled non-expert users to complete a physical hardware repair task without a traditional visual interface. The findings indicate that such architectures can support the sustainability of body-worn and ubiquitous technologies by enabling user-led maintenance through natural language interaction.

\balance

\bibliography{bibliography}

@article{Zhao02102023,
author = {Zhehui Zhao and Haoran Fu and Ruitao Tang and Bocheng Zhang and Yunmin Chen and Jianqun Jiang},
title = {Failure mechanisms in flexible electronics},
journal = {International Journal of Smart and Nano Materials},
volume = {14},
number = {4},
pages = {510--565},
year = {2023},
publisher = {Taylor \& Francis},
doi = {10.1080/19475411.2023.2261775},
URL = { 
    
        https://doi.org/10.1080/19475411.2023.2261775
},
eprint = { 
    
        https://doi.org/10.1080/19475411.2023.2261775
}
}

@inproceedings{10.1145/191666.191732,
author = {Mynatt, Elizabeth D. and Weber, Gerhard},
title = {Nonvisual presentation of graphical user interfaces: contrasting two approaches},
year = {1994},
isbn = {0897916506},
publisher = {Association for Computing Machinery},
address = {New York, NY, USA},
url = {https://doi.org/10.1145/191666.191732},
doi = {10.1145/191666.191732},
booktitle = {Proceedings of the SIGCHI Conference on Human Factors in Computing Systems},
pages = {166–172},
numpages = {7},
location = {Boston, Massachusetts, USA},
series = {CHI '94}
}

@article{Sweller1988CognitiveLD,
  title={Cognitive Load During Problem Solving: Effects on Learning},
  author={John Sweller},
  journal={Cogn. Sci.},
  year={1988},
  volume={12},
  pages={257-285}
}

@inproceedings{chulhong2015,
author = {Deterding, Sebastian and Lucero, Andrés and Holopainen, Jussi and Min, Chulhong and Cheok, Adrian and Waern, Annika and Walz, Steffen},
year = {2015},
month = {04},
pages = {2365-2368},
title = {Embarrassing Interactions},
doi = {10.1145/2702613.2702647}
}

@inproceedings{frederiksen2026goal,
  author    = {Frederiksen, Morten Roed},
  title     = {A Goal-Oriented Agentic Framework For Collaborative Branching Human-Robot Interactions},
  booktitle = {Proceedings of the 2026 IEEE Workshop on Advanced Robotics and its Social Impacts (ARSO)},
  year      = {2026},
  note      = {Under review}
}

@misc{chromadb_github,
  author = {Jeff Huber and Anton Troynikov},
  title = {Chroma},
  year = {2023},
  publisher = {GitHub},
  journal = {GitHub repository},
  howpublished = {\url{https://github.com/chroma-core/chroma}}
}

@article{Marge2020SpokenLI,
  title={Spoken Language Interaction with Robots: Research Issues and Recommendations, Report from the NSF Future Directions Workshop},
  author={Matthew Marge and Carol Y. Espy-Wilson and Nigel G. Ward and Abeer Alwan and Yoav Artzi and Mohit Bansal and Gil and Blankenship and Joyce Yue Chai and Hal Daum{\'e} and Debadeepta Dey and Mary P. Harper and Thomas M. Howard and Casey and Kennington and Ivana Kruijff-Korbayov{\'a} and Dinesh Manocha and Cynthia Matuszek and Ross Mead and Raymond and Mooney and Roger K. Moore and Marilyn Ostendorf and Heather Pon-Barry and Alex Rudnicky and Matthias and Scheutz and Robert St. Amant and Tong Sun and Stefanie Tellex and David R. Traum and Zhou Yu},
  journal={Comput. Speech Lang.},
  year={2020},
  volume={71},
  pages={101255}
}

@inproceedings{Brewster2003MultimodalI,
  title={Multimodal 'eyes-free' interaction techniques for wearable devices},
  author={Stephen Anthony Brewster and Joanna Lumsden and Marek Bell and Malcolm Hall and Stuart Tasker},
  booktitle={International Conference on Human Factors in Computing Systems},
  year={2003}
}

@inbook{Hancock07,
author = {Hancock, Peter and Szalma, James},
year = {2007},
month = {01},
pages = {195-206},
title = {Stress and Neuroergonomics},
isbn = {9780195177619},
journal = {Neuroergonomics: the Brain at Work},
doi = {10.1093/acprof:oso/9780195177619.003.0013}
}

@inproceedings{Profita2013DontMM,
  title={Don't mind me touching my wrist: a case study of interacting with on-body technology in public},
  author={Halley P. Profita and James Clawson and Scott M. Gilliland and Clint Zeagler and Thad Starner and Jim Budd and Ellen Yi-Luen Do},
  booktitle={International Semantic Web Conference},
  year={2013}
}

@article{Heikenfeld2018WearableSM,
  title={Wearable sensors: modalities, challenges, and prospects.},
  author={Jason C. Heikenfeld and Andrew J. Jajack and John A. Rogers and Philipp Gutruf and Limei Tian and Tingrui Pan and Ronald A. Li and Michelle Khine and Jayoung Kim and Joseph Wang},
  journal={Lab on a chip},
  year={2018},
  volume={18 2},
  pages={
          217-248
        }
}

@article{Henderson2011ExploringTB,
  title={Exploring the Benefits of Augmented Reality Documentation for Maintenance and Repair},
  author={Steven J. Henderson and Steven K. Feiner},
  journal={IEEE Transactions on Visualization and Computer Graphics},
  year={2011},
  volume={17},
  pages={1355-1368}
}

@article{buttussi21,
author = {Buttussi, Fabio and Chittaro, Luca},
year = {2021},
month = {02},
pages = {1-15},
title = {A Comparison of Procedural Safety Training in Three Conditions: Virtual Reality Headset, Smartphone, and Printed Materials},
volume = {14},
journal = {IEEE Transactions on Learning Technologies},
doi = {10.1109/TLT.2020.3033766}
}

@article{Kiesler2002MentalMO,
  title={Mental models of robotic assistants},
  author={Sara B. Kiesler and Jennifer Goetz},
  journal={CHI '02 Extended Abstracts on Human Factors in Computing Systems},
  year={2002}
}

@Article{s22197584,
AUTHOR = {Shi, Yongjun and Dong, Wei and Lin, Weiqi and Gao, Yongzhuo},
TITLE = {Soft Wearable Robots: Development Status and Technical Challenges},
JOURNAL = {Sensors},
VOLUME = {22},
YEAR = {2022},
NUMBER = {19},
ARTICLE-NUMBER = {7584},
URL = {https://www.mdpi.com/1424-8220/22/19/7584},
PubMedID = {36236683},
ISSN = {1424-8220},
DOI = {10.3390/s22197584}
}

@article{chen23,
author = {Chen, Lufeng and Xie, Hongqin and Liu, Zicheng and Li, Bin and Cheng, Hong},
year = {2023},
month = {12},
pages = {},
title = {Exploring Challenges and Opportunities of Wearable Robots: A Comprehensive Review of Design, Human-Robot Interaction and Control Strategy},
volume = {12},
journal = {APSIPA Transactions on Signal and Information Processing},
doi = {10.1561/116.00000156}
}

@article{Yao2022ReActSR,
  title={ReAct: Synergizing Reasoning and Acting in Language Models},
  author={Shunyu Yao and Jeffrey Zhao and Dian Yu and Nan Du and Izhak Shafran and Karthik Narasimhan and Yuan Cao},
  journal={ArXiv},
  year={2022},
  volume={abs/2210.03629}
}

@inproceedings{Ahn2022DoAI,
  title={Do As I Can, Not As I Say: Grounding Language in Robotic Affordances},
  author={Michael Ahn and Anthony Brohan and Noah Brown and Yevgen Chebotar and Omar Cortes and Byron David and Chelsea Finn and Keerthana Gopalakrishnan and Karol Hausman and Alexander Herzog and Daniel Ho and Jasmine Hsu and Julian Ibarz and Brian Ichter and Alex Irpan and Eric Jang and Rosario M Jauregui Ruano and Kyle Jeffrey and Sally Jesmonth and Nikhil Jayant Joshi and Ryan C. Julian and Dmitry Kalashnikov and Yuheng Kuang and Kuang-Huei Lee and Sergey Levine and Yao Lu and Linda Luu and Carolina Parada and Peter Pastor and Jornell Quiambao and Kanishka Rao and Jarek Rettinghouse and Diego M Reyes and Pierre Sermanet and Nicolas Sievers and Clayton Tan and Alexander Toshev and Vincent Vanhoucke and F. Xia and Ted Xiao and Peng Xu and Sichun Xu and Mengyuan Yan},
  booktitle={Conference on Robot Learning},
  year={2022}
}

@INPROCEEDINGS{794797,
  author={Peltola, J. and Plomp, J. and Seppanen, T.},
  booktitle={Proceedings 25th EUROMICRO Conference. Informatics: Theory and Practice for the New Millennium}, 
  title={A dictionary-adaptive speech driven user interface for a distributed multimedia platform}, 
  year={1999},
  volume={2},
  number={},
  pages={326-332 vol.2},
  doi={10.1109/EURMIC.1999.794797}
  }
\bibliographystyle{IEEEtran}

\end{document}